# Approximate inference on planar graphs using Loop Calculus and Belief Propagation

**Vicenç Gómez**
**Hilbert J. Kappen**
Radboud University Nijmegen,
Donders Institute for Brain, Cognition and Behaviour
6525 EZ Nijmegen, The Netherlands

**Michael Chertkov**
Theoretical Division and Center for Nonlinear Studies
Los Alamos National Laboratory
Los Alamos, NM 87545

## Abstract

We introduce novel results for approximate inference on planar graphical models using the loop calculus framework. The loop calculus (Chertkov and Chernyak, 2006b) allows to express the exact partition function $Z$ of a graphical model as a finite sum of terms that can be evaluated once the belief propagation (BP) solution is known. In general, full summation over all correction terms is intractable. We develop an algorithm for the approach presented in Chertkov et al. (2008) which represents an efficient truncation scheme on planar graphs and a new representation of the series in terms of Pfaffians of matrices. We analyze in detail both the loop series and the Pfaffian series for models with binary variables and pairwise interactions, and show that the first term of the Pfaffian series can provide very accurate approximations. The algorithm outperforms previous truncation schemes of the loop series and is competitive with other state-of-the-art methods for approximate inference.

## 1 Introduction

Graphical models are popular tools widely used in many areas which require modeling of uncertainty. They provide an effective approach through a compact representation of the joint probability distribution. The two most common types of graphical models are Bayesian Networks (BN) and Markov Random Fields (MRFs).

The partition function of a graphical model, which plays the role of normalization constant in a MRF or probability of evidence (likelihood) in a BN is a fundamental quantity which arises in many contexts such as hypothesis testing or parameter estimation. Exact computation of this quantity is only feasible when the graph is not too complex, or equivalently, when its tree-width is small. Currently many methods are devoted to approximate this quantity.

The belief propagation (BP) algorithm (Pearl, 1988) is at the core of many of these methods. Initially thought as an exact algorithm for tree graphs, it is widely used as an approximation method for loopy graphs (Murphy et al., 1999; Frey and MacKay, 1998). The exact partition function $Z$ is explicitly related to the BP approximation through the loop calculus framework introduced by Chertkov and Chernyak (2006b). Loop calculus allows to express $Z$ as a finite sum of terms (loop series) that can be evaluated once the BP solution is known. Each term maps uniquely to a subgraph, also denoted as a generalized loop, where the connectivity of any node within the subgraph is *at least* degree 2. Summation of the entire loop series is a hard combinatorial task since the number of generalized loops is typically exponential in the size of the graph. However, different approximations can be obtained by considering different subsets of generalized loops in the graph.

Although it has been shown empirically (Gómez et al., 2007; Chertkov and Chernyak, 2006a) that truncating this series may provide efficient corrections to the initial BP approximation, a formal characterization of the classes of tractable problems via loop calculus still remains as an open question. The work of Chertkov et al. (2008) represents a step in this direction, where it was shown that for any graphical model, summation of a certain subset of terms can be mapped to a summation of weighted perfect matchings on an extended graph. For planar graphs (graphs that can be embedded into a plane without crossing edges), this summation can be performed in polynomial time evaluating the Pfaffian of a skew-symmetric matrix associated with the extended graph. Furthermore, the full loop series can be expressed as a sum over Pfaffian terms, each one accounting for a large number of loops and solvable in polynomial time as well.

This approach builds on classical results from 1960s by Kasteleyn (1963); Fisher (1966) and other physicists who showed that in a *planar graphical model defined in terms of binary variables*, computing $Z$ can be mapped to a weighted perfect matching problem and calculated in polynomial time under the key restriction that interactions only depend on agreement or disagreement between the signs of their variables. Such a model is known in statistical physics as the Ising model *without external field*. Notice that exact inference for a general binary graphical model on a planar graph (i.e. Ising model with external field) is intractable (Barahona, 1982).

Recently, some methods for inference over graphical models, based on the works of Kasteleyn and Fisher, have been introduced. Globerson and Jaakkola (2007) obtained upper bounds on $Z$ for non-planar graphs with binary variables by decomposition of $Z$ into a weighted sum over partition functions of spanning tractable (zero field) planar models. Another example is the work of Schraudolph and Kamenetsky (2009) which provides a framework for exact inference on a restricted class of planar graphs using the approach of Kasteleyn and Fisher.

Contrary to the two aforementioned approaches which rely on exact inference on a tractable planar model, the loop calculus directly leads to a framework for approximate inference on general planar graphs. Truncating the loop series according to Chertkov et al. (2008) already gives the exact result in the zero external field case. In the general planar case, however, this truncation may result into an accurate approximation that can be incrementally corrected by considering subsequent terms in the series.

## 2  Belief Propagation and loop Series for Planar Graphs

We consider the Forney graph representation, also called general vertex model (Forney, 2001; Loeliger, 2004), of a probability distribution $p(\boldsymbol{\sigma})$ defined over a vector $\boldsymbol{\sigma}$ of binary variables (vectors are denoted using bold symbols). Forney graphs are associated with general graphical models which subsume other factor graphs, e.g. those correspondent to BNs and MRFs.

A binary Forney graph $\mathcal{G} := (\mathcal{V}, \mathcal{E})$ consists of a set of nodes $\mathcal{V}$ where each node $a \in \mathcal{V}$ represents an interaction and each edge $(a, b) \in \mathcal{E}$ represents a binary variable $ab$ which take values $\sigma_{ab} := \{\pm 1\}$. We denote $\bar{a}$ the set of neighbors of node $a$. Interactions $f_a(\boldsymbol{\sigma}_a)$ are arbitrary functions defined over typically small subsets of variables where $\boldsymbol{\sigma}_a$ is the vector of variables associated with node $a$, i.e. $\boldsymbol{\sigma}_a := (\sigma_{ab_1}, \sigma_{ab_2}, \ldots)$ where $b_i \in \bar{a}$. The joint probability distribution of such a model factorizes as:

$$p(\boldsymbol{\sigma}) = Z^{-1} \prod_{a \in \mathcal{V}} f_a(\boldsymbol{\sigma}_a), \quad Z = \sum_{\boldsymbol{\sigma}} \prod_{a \in \mathcal{V}} f_a(\boldsymbol{\sigma}_a), \quad (1)$$

where $Z$ is the partition function.

From a variational perspective, a fixed point of the BP algorithm represents a stationary point of the Bethe "free energy" approximation under proper constraints (Yedidia et al., 2005):

$$Z^{BP} = \exp\left(-F^{BP}\right), \quad (2)$$

$$F^{BP} = \sum_a \sum_{\boldsymbol{\sigma}_a} b_a(\boldsymbol{\sigma}_a) \ln\left(\frac{b_a(\boldsymbol{\sigma}_a)}{f_a(\boldsymbol{\sigma}_a)}\right)$$
$$- \sum_{b \in \bar{a}} \sum_{\sigma_{ab}} b_{ab}(\sigma_{ab}) \ln b_{ab}(\sigma_{ab}), \quad (3)$$

where $b_a(\boldsymbol{\sigma}_a)$ and $b_{ab}(\sigma_{ab})$ are the beliefs (pseudo-marginals) associated to each node $a \in \mathcal{V}$ and variable $ab$. For graphs without loops, Equation (2) coincides with the Gibbs "free energy" and therefore $Z^{BP}$ coincides with $Z$. If the graph contains loops, $Z^{BP}$ is just an approximation critically dependent on how strong the influence of the loops is. We introduce now some convenient definitions.

**Definition 1** *A **generalized loop** in a graph $\mathcal{G}$ is any subgraph $C$ such that each node in $C$ has degree 2 or larger.*

We use the term "loop" instead of "generalized loop" for the rest of the manuscript. $Z$ is explicitly represented in terms of the BP solution via the loop series expansion:

$$Z = Z^{BP} \cdot z, \quad z = \left(1 + \sum_{C \in \mathcal{C}} r_C\right), \quad r_C = \prod_{a \in C} \mu_{a; \bar{a}_C}, \quad (4)$$

where $\mathcal{C}$ is the set of all the loops within the graph. Each loop term $r_C$ is a product of terms $\mu_{a, \bar{a}_C}$ associated with every node $a$ of the loop. $\bar{a}_C$ denotes the set of neighbors of $a$ within the loop $C$:

$$\mu_{a; \bar{a}_C} = \frac{\displaystyle\sum_{\boldsymbol{\sigma}_a} b_a(\boldsymbol{\sigma}_a) \prod_{b \in \bar{a}_C} (\sigma_{ab} - m_{ab})}{\displaystyle\prod_{b \in \bar{a}_C} \sqrt{1 - m_{ab}^2}},$$

$$m_{ab} = \sum_{\sigma_{ab}} \sigma_{ab} b_{ab}(\sigma_{ab}). \quad (5)$$

We consider planar graphs with all nodes of degree not larger than 3, i.e. $|\bar{a}_C| \leq 3$. We denote by *triplet* a node with degree 3 in the graph.

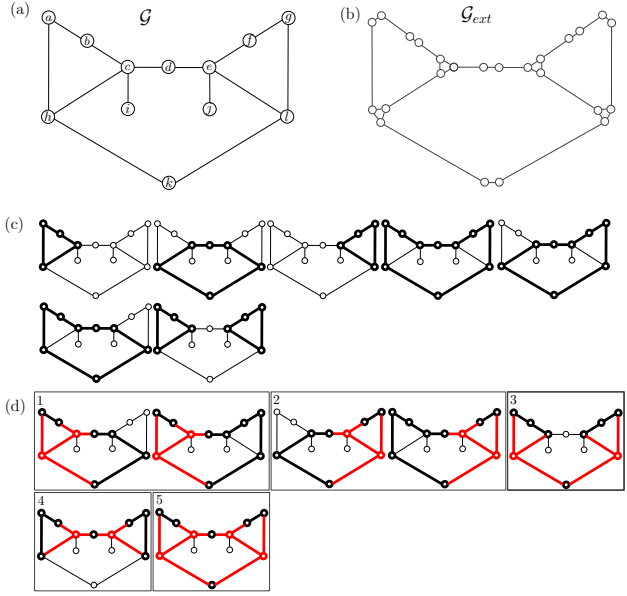

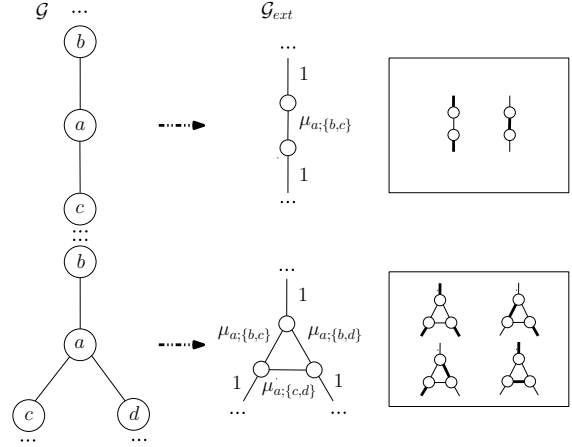

Figure 1: Example. **(a)** A Forney graph. **(b)** Corresponding extended graph. **(c)** Loops (in bold) included in the 2-regular partition function $Z_\emptyset$. **(d)** Loops (in bold and red) not included in $Z_\emptyset$. Marked in red, the triplets associated with each loop. Grouped in gray squares, the loops considered in different subsets $\Psi$ of triplets: (d.1) $\Psi = \{c, h\}$, (d.2) $\Psi = \{e, l\}$, (d.3) $\Psi = \{h, l\}$, (d.4) $\Psi = \{c, e\}$ and (d.4) $\Psi = \{c, e, h, l\}$ (see Section 2.2).

**Definition 2** *A **2-regular loop** is a loop in which all nodes have degree* exactly 2.

**Definition 3** *The **2-regular partition function** $Z_\emptyset$ is the truncated form of* (4) *which sums all 2-regular loops only:*

$$Z_\emptyset = Z^{BP} \cdot z_\emptyset, \quad z_\emptyset = 1 + \sum_{C \in \mathcal{C} \, s.t. \, |\bar{a}_C|=2, \forall a \in C} r_C. \quad (6)$$

Figures 1a and 1c show a small Forney graph and its seven 2-regular loops found in $Z_\emptyset$ respectively.

Finally, consider the set $P$ of all permutations $\alpha$ of the set $\mathcal{S} = \{1, \ldots, 2n\}$ in pairs: $\alpha = ((i_1, j_1), \ldots, (i_n, j_n))$, $i_k < j_k$, $\forall k = 1, \ldots, n$.

**Definition 4** *The **Pfaffian** of a skew-symmetric matrix $A = (A_{ij})_{1 \le i < j \le 2n}$ with $(A_{ij} = -A_{ji})$ is:*

$$\text{Pfaffian}(A) = \sum_{\alpha \in P} \text{sign}(\alpha) \prod_{(i,j) \in \alpha} A_{ij},$$

where the sign of a permutation $\alpha$ is $-1$ if the number of transpositions to get $\alpha$ from $\mathcal{S}$ is odd and $+1$ otherwise. The following identity allows to obtain the Pfaffian *up to a sign* by computing the determinant:

$$\text{Pfaffian}^2(A) = \text{Det}(A).$$

### 2.1 Computing the 2-regular Partition Function Using Perfect Matching

In Chertkov et al. (2008) it has been shown that computation of $Z_\emptyset$ in a Forney graph $\mathcal{G}$ can be mapped to the computation of the sum of all weighted perfect matchings within another extended weighted graph $\mathcal{G}_{ext} := (\mathcal{V}_{\mathcal{G}_{ext}}, \mathcal{E}_{\mathcal{G}_{ext}})$. A perfect matching is a subset of edges such that each node neighbors exactly one edge from the subset and its weight is the product of weights of edges in it. The key idea of this mapping is that each each perfect matching in $\mathcal{G}_{ext}$ corresponds to a 2-regular loop in $\mathcal{G}$. (See Figures 1b and c for an illustration). If $\mathcal{G}_{ext}$ is planar, the sum of all its weighted perfect matchings can be calculated in polynomial time evaluating the Pfaffian of an associated matrix. Here we reproduce these results with little variations and emphasis on the algorithmic aspects.

Given a Forney graph $\mathcal{G}$ and the BP approximation, we obtain the 2-core of $\mathcal{G}$ by removing nodes of degree 1 recursively. After this step, $\mathcal{G}$ is either the null graph (and then BP is exact) or it is only composed of vertices of degree 2 or 3.

To construct $\mathcal{G}_{ext}$ we split each node in $\mathcal{G}$ according to the rules introduced by Fisher (1966) and illustrated in Figure 2. Each 2-regular loop in $\mathcal{G}$ is associated with a perfect matching in $\mathcal{G}_{ext}$ and, furthermore, this correspondence is *unique*. Consider, for instance, the vertex $a$ of degree 3 in the bottom of Figure 2. Given a 2-regular loop $C$, $a$ can appear in four different configurations: either $a$ does not appear in $C$, or $C$ contains one of the following three paths: -b-a-c-, -b-a-d- or -c-a-d-. These four cases correspond to node terms in a loop with values 1, $\mu_{a;\{b,c\}}$, $\mu_{a;\{b,d\}}$ and $\mu_{a;\{c,d\}}$ respectively, and coincide with the matchings in $\mathcal{G}_{ext}$ shown within the box on the bottom-right. A simpler argument applies to the vertex of degree 2 of the top portion of Figure 2.

Figure 2: Fisher's rules. **(Top)** A node $a$ of degree 2 in $\mathcal{G}$ is split in 2 nodes in $\mathcal{G}_{ext}$. **(Bottom)** A node $a$ of degree 3 in $\mathcal{G}$ is split in 3 nodes in $\mathcal{G}_{ext}$. Right boxes include all matchings in $\mathcal{G}_{ext}$ related with node $a$ in $\mathcal{G}$.

Therefore, if we associate to each internal edge (new edge in $\mathcal{G}_{ext}$ not in $\mathcal{G}$) of each split node $a$ the corresponding term $\mu_{a;\bar{a}_C}$ of Equation (5) and to the external edges (existing edges already in $\mathcal{G}$) weight 1, then the sum over all weighted perfect matchings defined on $\mathcal{G}_{ext}$ is precisely $z_\emptyset$ (Equation 6).

Kasteleyn (1963) provided a method to compute this sum in polynomial time for planar graphs. First, edges are properly oriented in such a way that for every face (except possibly the external face) the number of clockwise oriented edges is odd. We use the linear time algorithm in Karpinski and Rytter (1998) to produce such an orientation.

Second, denote $\mu_{ij}$ the weight of the edge between nodes $i$ and $j$ in $\mathcal{G}'_{ext}$ and define the following skew-symmetric matrix:

$$\hat{A}_{ij} = \begin{cases} +\mu_{ij} & \text{if } (i,j) \in \mathcal{E}_{\mathcal{G}'_{ext}} \\ -\mu_{ij} & \text{if } (j,i) \in \mathcal{E}_{\mathcal{G}'_{ext}} \\ 0 & \text{otherwise} \end{cases}.$$

Calculation of $z_\emptyset$ can therefore be performed in time $\mathcal{O}(N^3_{\mathcal{G}_{ext}})$:

$$z_\emptyset = \sqrt{\text{Det}(\hat{A})}.$$

For the special case of binary planar graphs with zero local fields (Ising model *without external field*) we have $Z = Z_\emptyset = Z^{BP} \cdot z_\emptyset$ since the other terms in the loop series vanish.

### 2.2 Computing the Full Loop Series Using Perfect Matching

Chertkov et al. (2008) established that $z_\emptyset$ is just the first term of a finite sum involving Pfaffians. We briefly reproduce their results here and provide an algorithm to compute the full loop series as a Pfaffian series.

Consider $\mathcal{T}$ defined as the set of all possible triplets (vertices with degree 3 in the original graph $\mathcal{G}$). For each possible subset $\Psi \in \mathcal{T}$, including an *even* number of triplets, there exists a unique correspondence between loops in $\mathcal{G}$ including the triplets in $\Psi$ and perfect matchings in another extended graph $\mathcal{G}_{ext_\Psi}$ constructed after removal of the triplets $\Psi$ in $\mathcal{G}$. Using this representation the full loop series can be represented as a Pfaffian series, where each term $Z_\Psi$ is tractable and is a product of respective Pfaffians and $\mu_{a;\bar{a}}$ terms associated with each triplet of $\Psi$:

$$z = \sum_\Psi Z_\Psi, \qquad Z_\Psi = z_\Psi \prod_{a \in \Psi} \mu_{a;\bar{a}}$$

$$z_\Psi = \text{sign}\left(\text{Pfaffian}\left(\hat{B}_\Psi\right)\right) \cdot \text{Pfaffian}\left(\hat{A}_\Psi\right), \quad (7)$$

where $\hat{B}_\Psi$ corresponds to the original Kasteleyn matrix with weights $+1$ instead of $+\mu_{ij}$ and $-1$ instead of $-\mu_{ij}$ and we make explicit use of the Pfaffians to correct for the sign.[1]

Note that the 2-regular partition function thus corresponds to the particular case $\Psi = \emptyset$. We refer to the remaining terms of the series ($\Psi \neq \emptyset$) as higher order terms. Notice that matrices $\hat{A}_\Psi$ and $\hat{B}_\Psi$ depend on the removed triplets and therefore each $z_\Psi$ requires different matrices and different edge orientations. Figure 1d shows loops corresponding to the higher order terms on our illustrative example.

Exhaustive enumeration of all subsets of triplets leads to a prohibitive $2^{|\mathcal{T}|}$ time algorithm. However, the key advantage of the Pfaffian representation is that $Z_\emptyset$ is always the term that accounts for the largest number of loop terms in the series. Algorithm 1 describes how to compute all Pfaffian terms. It can be interrupted at any time, leading to incremental corrections.

---
**Algorithm 1** Pfaffian series
---
**Require:** Forney graph $\mathcal{G}$
1: $z := 0$.
2: **for all** ($\Psi \in \mathcal{T}$) **do**
3:   Build $\mathcal{G}_{ext_\Psi}$ applying rules of Figure 2.
4:   Set Pfaffian orientation in $\mathcal{G}_{ext_\Psi}$.
5:   Build matrices $\hat{A}$ and $\hat{B}$.
6:   Compute $z_\Psi$ according to Equation (7).
7:   $z := z + z_\Psi \prod_{a \in \Psi} \mu_{a;\bar{a}}$.
8: **end for**
9: **RETURN** $Z^{BP} \cdot z$
---

## 3 Experiments

In this Section we study numerically the proposed algorithm. We focus on the binary Ising model, a particular case of (1) where factors only depend on the disagreement between two variables and take the form $f_a(\sigma_{ab}, \sigma_{ac}) = \exp\left(J_{a;\{ab,ac\}} \sigma_{ab} \sigma_{ac}\right)$. We introduce nonzero local potentials $f_a(\sigma_{ab}) = \exp(J_{a;\{ab\}}\sigma_{ab})$ so that the model becomes planar-intractable. We create different inference problems by choosing normally distributed interactions and local field parameters, i.e. $\{J_{a;\{ab,ac\}}\} \sim \mathcal{N}(0, \beta/2)$ and $\{J_{a;\{ab\}}\} \sim \mathcal{N}(0, \beta\Theta)$. $\Theta$ and $\beta$ determine how difficult the inference problem is. Generally, for $\Theta = 0$ the planar problem is tractable. For $\Theta > 0$, small values of $\beta$ result in weakly coupled variables (easy problems) and large values of $\beta$ in strongly coupled variables (hard problems). Larger values of $\Theta$ result in easier inference problems.

---
[1] The correction sign(Pfaffian($\hat{B}_\Psi$)) is necessary because loop terms can be negative and even evaluating Pfaffian($\hat{A}_\Psi$) with the correct sign would only give the contribution up to a sign. In the previous subsection, we assumed $z_\emptyset$ positive.

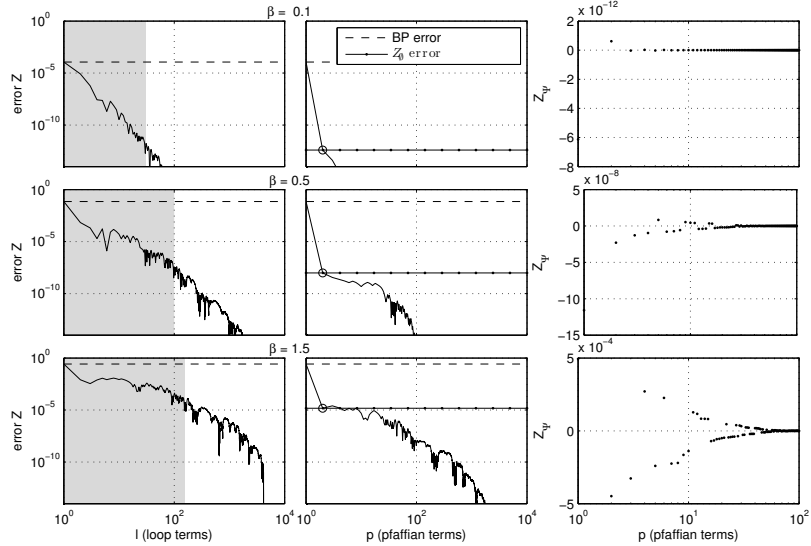

Figure 3: Comparison between the loop series and the Pfaffian series. Each row corresponds to a different value of the interaction strength $\beta$. **Left column** shows the $Z^{TLSBP}(l)$ error in log-log scale. Shaded regions include all loop terms (not necessarily 2-regular loops) required to reach the same (or better) accuracy than $Z_\emptyset$. **Middle column** shows the $Z^{Pf}(p)$ error. The first Pfaffian term corresponds to $Z_\emptyset$, marked by a circle. **Right column** shows the values of the first 100 higher order terms sorted in descending order in $|Z_\Psi|$ and excluding $z_\emptyset$.

To evaluate the approximations we consider errors in $Z$ and, when possible, computational cost as well. As shown in Gómez et al. (2007), errors in $Z$ are equivalent to errors in single variable marginals, which can be obtained by conditioning over the variables under interest. We consider tractable instances via the junction tree algorithm (Lauritzen and Spiegelhalter, 1988) using 8GB of memory. Given an approximation $Z'$ of $Z$, we use $\frac{|\log Z - \log Z'|}{\log Z}$ as error measure.

### 3.1 Full Pfaffian Series

Here we analyze numerically how the loop and Pfaffian representations differ using an example, shown in Figure 4 as a factor graph, for which all terms of both series can be computed.

We use TLSBP (Gómez et al., 2007) to retrieve all loops (8123 for this example) and Algorithm 1 to com-pute all Pfaffian terms. To compare both approximations we sort all contributions, either loops or Pfaffians, by their absolute values in descending order, and then analyze how errors are corrected as more terms are included in the respective series. We define partition functions for the truncated series as:

$$Z^{TLSBP}(l) = Z^{BP}\left(1 + \sum_{i=1\ldots l} r_{C_i}\right), \quad (8)$$
$$Z^{Pf}(p) = Z^{BP}\left(\sum_{i=1\ldots p} Z_{\Psi_i}\right).$$

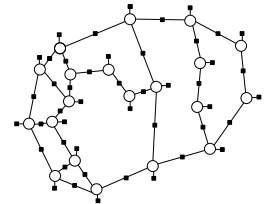

Figure 4: Planar bipartite factor graph used to compare the full Pfaffian series with the loop series. Circles and black squares denote variables and factors respectively. We use $\Theta = 0.1$ and $\beta \in \{0.1, 0.5, 1.5\}$.

Then $Z^{TLSBP}(l)$ accounts for the $l$ most contributing loops and $Z^{Pf}(p)$ for the $p$ most contributing Pfaffian terms. In all cases, $Z_{\Psi_1}$ corresponds to $z_\emptyset$.

Figure 3 shows the error $Z^{TLSBP}$ (first column) and $Z^{Pf}$ (second column) for both representations. For weak interactions ($\beta = 0.1$) BP converges fast and provides an accurate approximation with an error of order $10^{-4}$. Summation of less than 50 loop terms (top-left panel) is enough to obtain machine precision accuracy. The BP error is almost reduced totally with the $z_\emptyset$ correction (top-middle panel). In this scenario, higher order terms are negligible (top-right panel).

For strong interactions ($\beta = 1.5$) BP converges after many iterations and gives a poor approximation. Also a larger proportion of loop terms (bottom-left panel) is necessary to correct the BP result up to machine precision. Looking at the bottom-left panel we find that approximately 200 loop terms are required to achieve the same correction as the one obtained by $z_\emptyset$. The $z_\emptyset$ is quite accurate (bottom-middle panel).

As the right panels show, as $\beta$ increases, higher order terms change from a flat sequence of small terms to an alternating sequence of positive and negative terms which grow in absolute value. These oscillations are also present in the loop series representation.

We conclude that the $z_\emptyset$ correction can give a significant improvement even in hard problems for which BP converges after many iterations. Notice again that calculating $z_\emptyset$ does not require explicit search of loop or Pfaffian terms.

### 3.2 Square Grids

We now analyze the $Z_\emptyset$ approximation using Ising grids (nearest neighbor connectivity) and compare with the following methods: [2]

**Truncated Loop-Series for BP** (TLSBP) (Gómez et al., 2007), which accounts for a certain number of loops by performing depth-first-search on the graph and then merging found loops iteratively. We adapted TSLBP as an any-time algorithm (**anyTLSBP**) such that only the length of the loop $l$ is used as parameter. We run anyTLSBP by selecting loops shorter than a given $l$, and increased $l$ progressively.

**Cluster Variation Method** (**CVM-Loopk**) A double loop implementation of CVM (Heskes et al., 2003), which is a special case of generalized belief propagation (Yedidia et al., 2005) with convergence guarantees. We use as outer clusters all (maximal) factors together with loops of four (k=4) or six (k=6) variables in the factor graph.

**Tree-Structured Expectation Propagation** (**TreeEP**) (Minka and Qi, 2004). This method performs exact inference on a base tree of the graphical model and approximates the other interactions.

We also compare with two variational methods which give upper bounds on $Z$: **Tree Reweighting** (**TRW**)

---

[2] We use the libDAI library (Mooij, 2008) for algorithms **CVM-Loopk**, **TreeEP** and **TRW**.

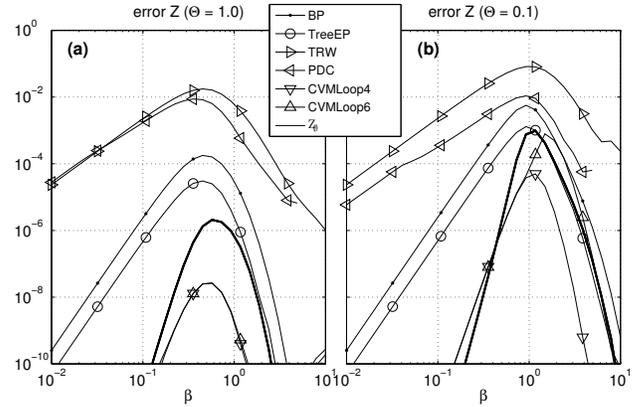

Figure 5: 7x7 grid **attractive interactions** and positive local fields. BP converged always. **Error averages** over 50 random instances as a function of $\beta$ for **(a)** strong local fields and **(b)** weak local fields.

(Wainwright et al., 2005), which decomposes the parametrization of a probabilistic graphical model as a mixture of spanning trees, and **Planar graph decomposition** (**PDC**) (Globerson and Jaakkola, 2007), which uses a mixture of *planar-tractable* graphs.

#### 3.2.1 Attractive Interactions

We first focus on models with interactions that tend to align neighboring variables to the same value, $J_{a;\{ab,ac\}} > 0$. If local fields are also positive, $J_{a;\{ab\}} > 0, \forall a \in \mathcal{V}$, Sudderth et al. (2008) showed that $Z^{BP}$ is a *lower bound* of $Z$ and all loop terms (and therefore Pfaffian terms too) have the same sign.

We generate grids with $|\{J_{a;bc}\}| \sim \mathcal{N}(0, \beta/2)$ and $|\{J_{a;b}\}| \sim \mathcal{N}(0, \beta\Theta)$ for different interaction strengths and strong/weak local fields. Figure 5 shows average errors. All methods show an initial growth and a subsequent decrease, a fact explained by the phase transition occurring in this model for $\Theta = 0$ and $\beta \approx 1$ (Mooij and Kappen, 2005). As the difference between the two plots suggest, errors are larger as $\Theta$ approaches zero. Notice that $Z_\emptyset = Z$ for the limit case $\Theta = 0$, suggesting an abrupt change in the difficulty of the inference task from $\Theta = 0$ to $\Theta > 0$.

$Z_\emptyset$ *always improves* over the BP approximation. Corrections are most significant for weak interactions $\beta < 1$ and strong local fields.

The $Z_\emptyset$ approximation performs better than TreeEP in all cases except for very strong couplings, where they show very similar results. Interestingly, for $\Theta = 0.1$, $Z_\emptyset$ performs very similar to CVM which is known to be a very accurate approximation in grids. Notice that using larger outer-clusters in CVM does not necessar-ily lead to improvements. Finally, PDC performs better than TRW [3]. In general, both upper bounds are significantly less tight than the lower bounds provided by BP and $Z_\emptyset$.

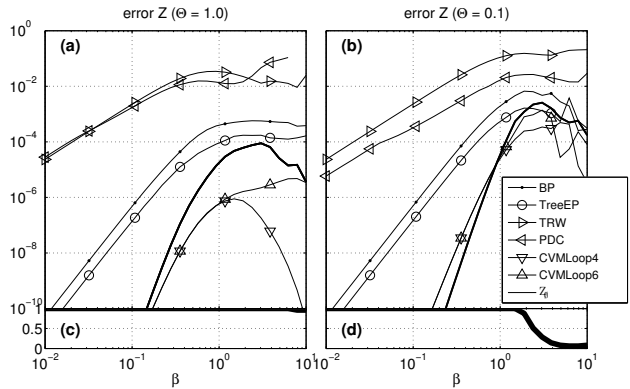

Figure 6: 7x7 grid **mixed interactions**. **Error averaged** over 50 random instances as a function of $\beta$ for **(a)** strong and **(b)** weak local fields. Bottom panels, **(c)** and **(d)**, show **proportion** of instances for which BP converged.

### 3.2.2 Mixed Interactions

Here we focus on a more general model where interactions and local fields can have mixed signs, $Z_\emptyset$ is no longer a lower bound and loop terms can be positive or negative. Figure 6 shows results using this setup.

For strong local fields (subplots a,c), we observe that $Z_\emptyset$ improvements become less significant as $\beta$ increases. It is important to note that $Z_\emptyset$ always improves on the BP result, even when the couplings are very strong ($\beta = 10$) and BP fails to converge in some instances. $Z_\emptyset$ performs substantially better than TreeEP for small and intermediate $\beta$. As in the case of attractive interactions, the best results are attained using CVM.

For the case of weak local fields (subplots b,d), BP fails to converge near the transition to the spin-glass phase. For $\beta > 2$, all methods give results comparable to BP or worse. For $\beta = 10$, BP converges only in a few instances, and it may happen that $Z_\emptyset$ degrades the $Z^{BP}$ approximation because loops of alternating signs have major influence in the series.

### 3.2.3 Scaling with Graph Size

We now study how the accuracy of $Z_\emptyset$ changes as a function of the grid size for $\sqrt{N} = \{4, ..., 18\}$ focusing on a difficult regime according to the previous results.

---

[3] The worse performance of PDC for strong couplings might be attributed to implementation artifacts.

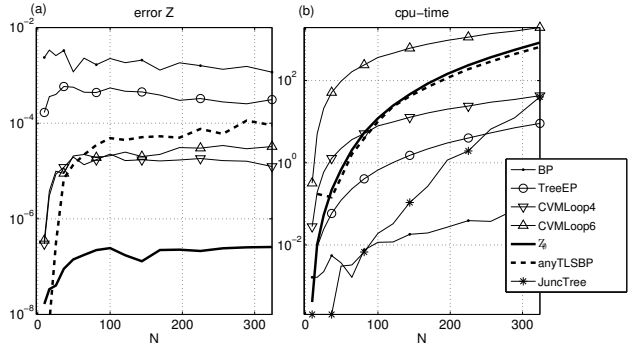

Figure 7: Scaling results on grids as a function of the network size for **strong** interactions $\beta = 1$ and **very weak** local fields $\Theta = 0.01$. BP converged in all cases. **(a) Error medians** over 50 instances. **(b)** Cpu time.

We compare also with anyTLSBP, which we run for the same cpu time as $Z_\emptyset$.

Figure 7a shows approximation errors. In increasing accuracy we get BP, TreeEP, anyTLSBP, CVM-Loop6, CVM-Loop4 and $Z_\emptyset$ which turns out to be the best approximation for this setup.

Overall, results are roughly independent of the network size $N$ in almost all methods that we compare. The error of anyTLSBP starts being the smallest but soon increases because the proportion of loops captured decreases very fast. The $Z_\emptyset$ correction, on the other hand, stays flat and it scales reasonably well. Interestingly, although $Z_\emptyset$ and anyTLSBP use different ways to truncate the loop series, both methods show similar scaling behavior for large graphs.

Figure 7b shows averaged cpu time for all tested algorithms. Although the cpu time required to compute $Z_\emptyset$ scales with $O(N_{G_{ext}}^3)$, its curve shows the steepest growth. The cpu time of the junction tree quickly increases with the tree-width of the graphs. For large enough $N$, exact solution via the junction tree method is no longer feasible because of its memory requirements. In contrast, for all approximate inference methods, memory demands do not represent a limitation.

## 4 Discussion

We have presented an approximate algorithm for the partition function based on the exact loop calculus framework for inference on planar graphical models defined in terms of binary variables. The proposed approach improves the estimate provided by BP without an explicit search of loops and turns out to be competitive with other state of the art methods.

Currently, the shortcoming of the presented approach is in its relatively costly implementation. However, since the bottleneck of the algorithm is the Pfaffian calculation and not the algorithm itself (used to obtain the extended graphs and the associated matrices), it is easy to devise more efficient methods than the one used here. Thus, one may substitute brute-force evaluation of the Pfaffians by a smarter one, which could reduce the cost from $\mathcal{O}(N^3_{G_{ext}})$ to $\mathcal{O}(N^{3/2}_{G_{ext}})$ (Loh and Carlson, 2006). This is focus of current investigation.

For direct comparison with other methods, we have focused on inference problems defined on planar graphs with symmetric pairwise interactions and, to make the problems difficult, we have introduced local field potentials. Notice however, that the algorithm can also be used to solve models with more complex interactions, i.e. more than pairwise as in the case of the Ising model (see Chertkov et al., 2008, for a discussion of possible generalization). This makes our approach more general than the approaches of Globerson and Jaakkola (2007); Schraudolph and Kamenetsky (2009), designed specifically for the pairwise interaction case.

Although planarity is a severe restriction, planar graphs appear in many contexts such as computer vision and image processing, magnetic and optical recording, or network routing and logistics. We are currently investigating possible extensions for approximate inference on some class of non-planar graphs for which efficient inference could be done in the "closest" planar graph.

### Acknowledgements

We acknowledge J. M. Mooij, A. Windsor and A. Globerson for providing their software and V. Y. Chernyak, J. K. Johnson and N. Schraudolph for interesting discussions. This research is part of the Interactive Collaborative Information Systems (ICIS) project, supported by the Dutch Ministry of Economic Affairs, grant BSIK03024. The work at LANL was carried out under the auspices of the National Nuclear Security Administration of the U.S. Department of Energy at Los Alamos National Laboratory under Contract No. DE-AC52-06NA25396.

### References

Barahona, F. (1982). On the computational complexity of Ising spin glass models. *J. Phys. A-Math. Gen.*, 15(10):3241–3253.

Chertkov, M. and Chernyak, V. Y. (2006a). Loop calculus helps to improve Belief Propagation and linear programming decodings of LDPC codes. In *invited talk at 44th Allerton Conference*.

Chertkov, M. and Chernyak, V. Y. (2006b). Loop series for discrete statistical models on graphs. *J. Stat. Mech-Theory E.*, 2006(06):P06009.

Chertkov, M., Chernyak, V. Y., and Teodorescu, R. (2008). Belief propagation and loop series on planar graphs. *J. Stat. Mech-Theory E.*, 2008(05):P05003 (19pp).

Fisher, M. (1966). On the dimer solution of the planar Ising model. *J. Math. Phys.*, 7(10):1776–1781.

Forney, G.D., J. (2001). Codes on graphs: normal realizations. *IEEE T. Inform. Theory*, 47(2):520–548.

Frey, B. J. and MacKay, D. J. C. (1998). A revolution: belief propagation in graphs with cycles. In *NIPS 10*, pages 479–486.

Globerson, A. and Jaakkola, T. S. (2007). Approximate inference using planar graph decomposition. In *NIPS 19*, pages 473–480.

Gómez, V., M. Mooij, J., and J. Kappen, H. (2007). Truncating the loop series expansion for belief propagation. *J. Mach. Learn. Res.*, 8:1987–2016.

Heskes, T., Albers, K., and Kappen, H. J. (2003). Approximate inference and constrained optimization. In *19th UAI*, pages 313–320.

Karpinski, M. and Rytter, W. (1998). Fast parallel algorithms for graph matching problems. pages 164–170. Oxford University Press, USA.

Kasteleyn, P. W. (1963). Dimer statistics and phase transitions. *J. Math. Phys.*, 4(2):287–293.

Lauritzen, S. L. and Spiegelhalter, D. J. (1988). Local computations with probabilities on graphical structures and their application to expert systems. *J. Roy. Stat. Soc. B Met.*, 50(2):154–227.

Loeliger, H.-A. (2004). An introduction to factor graphs. *IEEE Signal Proc. Mag.*, 21(1):28–41.

Loh, Y. L. and Carlson, E. W. (2006). Efficient algorithm for random-bond Ising models in 2d. *Phys. Rev. Lett.*, 97(22):227205.

Minka, T. and Qi, Y. (2004). Tree-structured approximations by expectation propagation. In *NIPS 16*.

Mooij, J. M. (2008). libDAI: A free/open source C++ library for discrete approximate inference methods.

Mooij, J. M. and Kappen, H. J. (2005). On the properties of the Bethe approximation and loopy belief propagation on binary networks. *J. Stat. Mech-Theory E.*, 2005(11):P11012.

Murphy, K. P., Weiss, Y., and Jordan, M. I. (1999). Loopy Belief Propagation for approximate inference: An empirical study. In *15th UAI*, pages 467–475.

Pearl, J. (1988). *Probabilistic Reasoning in Intelligent Systems: Networks of Plausible Inference*. Morgan Kaufmann Publishers, San Francisco, CA.

Schraudolph, N. and Kamenetsky, D. (2009). Efficient exact inference in planar Ising models. In *NIPS 21*.

Sudderth, E., Wainwright, M., and Willsky, A. (2008). Loop series and Bethe variational bounds in attractive graphical models. In *NIPS 20*, pages 1425–1432.

Wainwright, M., Jaakkola, T., and Willsky, A. (2005). A new class of upper bounds on the log partition function. *IEEE T. Inf. Theory*, 51(7):2313–2335.

Yedidia, J. S., Freeman, W. T., Weiss, Y., and Yuille, A. L. (2005). Constructing free-energy approximations and generalized belief propagation algorithms. *IEEE T. Inf. Theory*, 51(7):2282–2312.